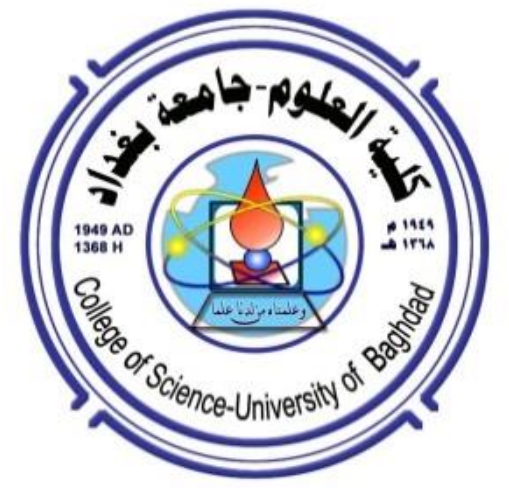

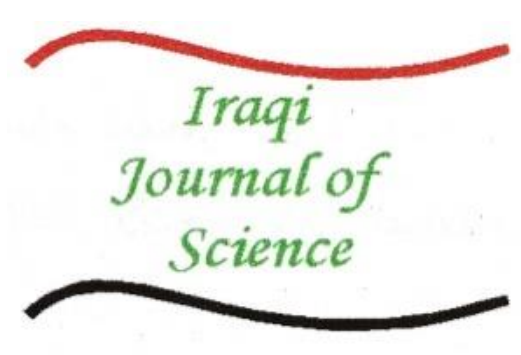

# Transfer Learning-Based Real-Time Handgun Detection

**Youssef Elmir [1,2]*, Abdeldjalil Abdelaziz[2], Mohammed Haidas[2]**

[1]*Laboratoire LITAN, École supérieure en Sciences et Technologies de l'Informatique et du Numérique RN 75, Amizour 06300, Bejaia, Algérie.*

[2]*SGRE-lab, University Tahri Mohammed of Bechar, Bechar, Algeria*

**Abstract**

Traditional surveillance systems rely on human attention, limiting their effectiveness. This study employs convolutional neural networks and transfer learning to develop a real-time computer vision system for automatic handgun detection. A comprehensive analysis of online handgun detection methods is conducted, emphasizing reducing false positives and learning time. Transfer learning is demonstrated as an effective approach. Despite technical challenges, the proposed system achieves a precision rate of 84.74%, demonstrating promising performance comparable to related works, enabling faster learning and accurate automatic handgun detection for enhanced security. This research advances security measures by reducing human monitoring dependence, showcasing the potential of transfer learning-based approaches for efficient and reliable handgun detection.

## 1. Introduction

Gun violence has become an escalating concern in the United States, with a significant increase in mass shooting incidents over recent years. In 2015 alone, the Mass Shooting Tracker recorded 372 such incidents, resulting in the tragic loss of 475 lives and injuries to 1,870 individuals [1]. Alongside the rise in gun-related crimes, the incidence of firearm thefts has also surged. In 2015, the FBI reported a total of 4,091 bank robberies, with 1,725 involving the use of firearms [2]. Existing security measures, such as silent alarms and panic buttons, rely on individuals' ability to activate them manually, a task that becomes daunting when confronted by an armed threat.

The Portland Police Bureau [3] advises individuals to remain composed and comply with robbers' instructions as a flight response strategy. However, this approach has inherent challenges. Users may struggle to activate the button, potentially delaying the alert to authorities or rendering it ineffective. Even when activated, the response time may allow criminals to escape or result in unfortunate accidents. In public areas like parks or other establishments, installing signaling devices can be logistically challenging. Furthermore, in locations lacking such alert systems, individuals attempting to seek help through alternative means may encounter obstacles. Addressing these limitations and identifying opportunities for improvement in surveillance and control systems are critical for safeguarding lives and preventing theft, necessitating enhancements in effectiveness and response time during emergencies.

*Email: elmir@estin.dz

In today's digital era, computer technology has revolutionized data management, enabling seamless storage, processing, indexing, and retrieval of information. Significant strides have been made in object detection, thanks to extensive research efforts and the availability of international image databases for machine learning. These databases have played a pivotal role in advancing methodologies in this field.

The increasing reliance on digital images, motion analysis, and object detection in videos underscores their pivotal role in various applications, particularly in video surveillance. Public safety concerns are paramount in modern society and demand attention. Weapons, especially in densely populated areas, pose significant threats to individuals' safety and security. Thus, relying solely on human operators may prove insufficient for preventing perilous situations during major events. This raises the question of harnessing artificial intelligence-based systems to provide comprehensive security solutions.

The utilization of online automatic handgun detection has the potential to improve the effectiveness of surveillance methods through the application of deep learning. Deep learning techniques, like recurrent neural networks, have been shown to be better than older methods like naive Bayes and decision trees in recent years. These techniques are better at many tasks, such as image classification, detection, and segmentation [4, 5, 6, 7, 8, 9]. Additionally, similar research has been suggested in [10], where an optimal detector demonstrates reliable results as an automated alarm system. However, to train deep convolutional neural networks (CNNs), which have millions of parameters, you need large datasets with millions of samples and access to high-performance computing (HPC) resources, like systems with multiple processors and graphics processing unit (GPU) acceleration.
Extensive research has been dedicated to developing firearms detection and prevention technologies, such as X-rays, electromagnetic scintigraphy, and profiling, to identify firearms in individuals. However, these technologies face limitations, including potential infringement on Second Amendment rights, leading to false alarms. To address this, emerging technologies like computer vision (CV) offer promise. CV systems analyze video streams, extracting contextual information to detect potential threats effectively. Intelligent video surveillance (IVS) systems leveraging CV can autonomously monitor areas without human intervention, enhancing firearm detection while considering context. These systems follow a pattern of image processing, tracking, and behavior analysis, providing comprehensive surveillance capabilities [7].

In an Intelligent Video Surveillance (IVS) system, video input undergoes several steps: Background model generation identifies foreground pixels, grouping them into targets (blobs) for activity monitoring. This process efficiently filters CCTV system data, providing relevant video metadata. High-level algorithms process this metadata, detecting predefined events and triggering responses, such as alarms. IVS systems, with real-time processing capabilities, can monitor multiple cameras simultaneously, potentially replacing human agents in CCTV surveillance. This typical structure in automated surveillance systems involves image processing, tracking, and higher-level scene and behavior analysis modules [7]. The process is visually illustrated in Figure 1.

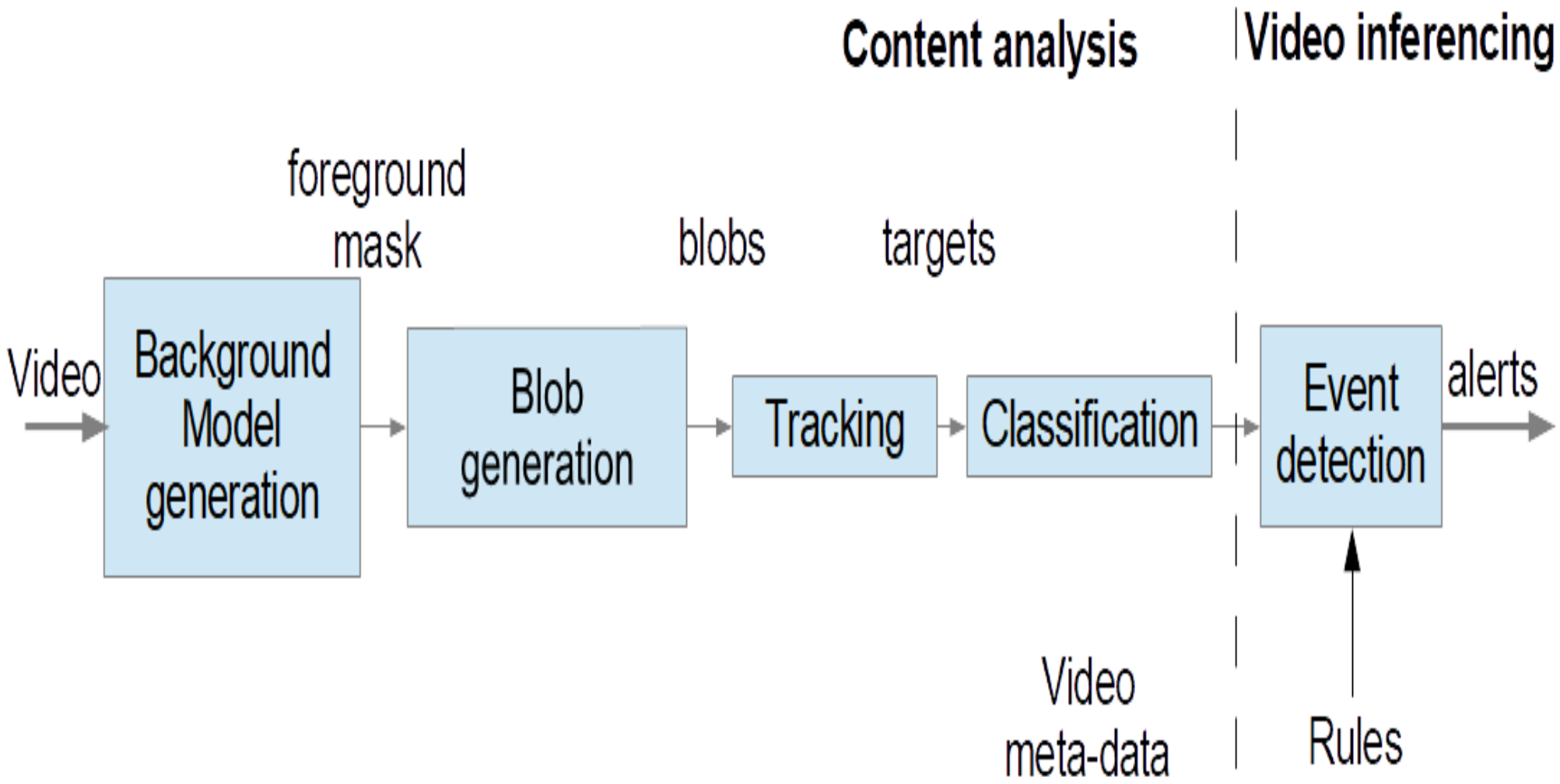

**Figure 1:** Organizational chart of the computer vision pipeline designed for security applications [11]

The potential of computer vision to enhance safety is acknowledged. Threatening poses are identified, and there is a promise of prompt alerting by authorities, with an 80% accuracy rate. Despite challenges, security and public safety can be significantly enhanced by it [12].
Researchers at AGH University in Krakow, Poland [13] used computer vision to detect firearms and knives in CCTV footage. They achieved 94.93% specificity and 81.18% sensitivity for knife detection and 96.69% specificity and 35.98% sensitivity for firearms. While more improvements can be made, this research highlights the potential of computer vision to enhance security through advanced surveillance systems [13].
Firearms were found to be detected more accurately than knives in [14], primarily due to limited database diversity and low-resolution CCTV recordings. To address this issue, the implementation of higher-quality cameras for faster threat detection is planned. Additionally, terahertz detection initiatives in the 100–500 GHz range are being pursued with the aim of enhancing security through the detection of concealed dangerous items.
In this concise overview of related works, the potential of computer vision in safety enhancement and the successful detection of firearms and knives in surveillance footage are emphasized. These insights underscore the significance of accuracy and privacy in weapon detection systems (see Table 1).

**Table 1:** Summary of Related Works

| Related Works | Accuracy (%) | Specificity (%) | Sensitivity (%) |
|---|---|---|---|
| Identifying threatening poses [12] | 80 | | |
| AGH University (Knives Detection) [13] | | 94.93 | 81.18 |
| AGH University (Firearms Detection) [13] | | 96.69 | 35.98 |

Privacy is prioritized in this work, and invasive methods for identifying weapons are avoided. Alerts are only triggered when weapons are openly displayed or used, ensuring safety without unnecessary police notifications for concealed carry.

This work shares similarities with a previous project [13], both aiming to automate weapon detection to address information overload in CCTV systems. However, the previous project faced limitations, particularly in firearm detection, due to a high number of false positives. To overcome these challenges, the goal is to enhance the system through specialized hardware utilization, incorporating hardware preprocessing for faster and more consistent weapon detection results. Additionally, transfer learning is employed to improve efficiency, leveraging pre-trained models' knowledge to accelerate learning and enhance detection accuracy, particularly with limited training data. This research demonstrates the effectiveness of transfer learning in developing a fast and accurate handgun detection system, contributing to computer vision's advancement in public safety and security.

The subsequent sections of this paper will cover various aspects. Section 2 will delve into the detailed explanation of the proposed methodology, while Section 3 will describe the experiments conducted and the datasets employed. The results obtained will be presented and analyzed in Section 4. Finally, Section 5 will summarize and conclude this work.

## 2. Proposed methodology

The task of handgun detection involves recognizing objects and determining their positions in static images or video sequences. In this study, we propose and test a basic CNN model (shown in Figure 2) for finding handguns using the same datasets and both the sliding window method and the region proposal method [10].

The process begins with image acquisition, involving two stages: capturing the initial images and capturing subsequent images. The subsequent images are essential for motion detection, as this process cannot be performed without them. The quality of the results relies heavily on the careful acquisition of images, emphasizing the importance of this step in achieving successful registration.

### *2.1 Motion Detection*

In contrast to static images, motion detection is crucial for handgun detection within a video sequence. However, due to its computational cost, it is important to optimize the process. To prevent unnecessary triggering of the handgun detection algorithm in the absence of motion in the video, a set of simple image processing operations is performed to identify the presence of a moving object, where applicable. This helps improve the efficiency and accuracy of the overall detection process [8].

Differential images [15] are obtained by subtracting two images:

$$\mathrm{Gdif}(x,y) = g1(x,y) - g2(x,y) \tag{1}$$

A differential image *Gdif* highlights motion by revealing variations between two images *g1* and *g2*. It involves computing the differences between three consecutive images: $I_{t-1}$, $I_t$, and $I_{t+1}$. This method offers the advantage of removing static background information and concentrating solely on the relevant changes associated with motion detection:

$$\Delta I_1 = I_{t+1} - I_t, \Delta I_2 = I_t - I_{t-1}, \Delta I = \Delta I_1 \wedge \Delta I_2 \tag{2}$$

In practical implementation, three images are captured at times t-1, t, and t+1 to compute the differences $\Delta I_1$ and $\Delta I_2$. $\Delta I_1$ represents the absolute difference between the last two images, while $\Delta I_2$ represents the absolute difference between the first two images. The final differential image, $\Delta I$, is obtained by taking the bitwise difference between $\Delta I_1$ and $\Delta I_2$.

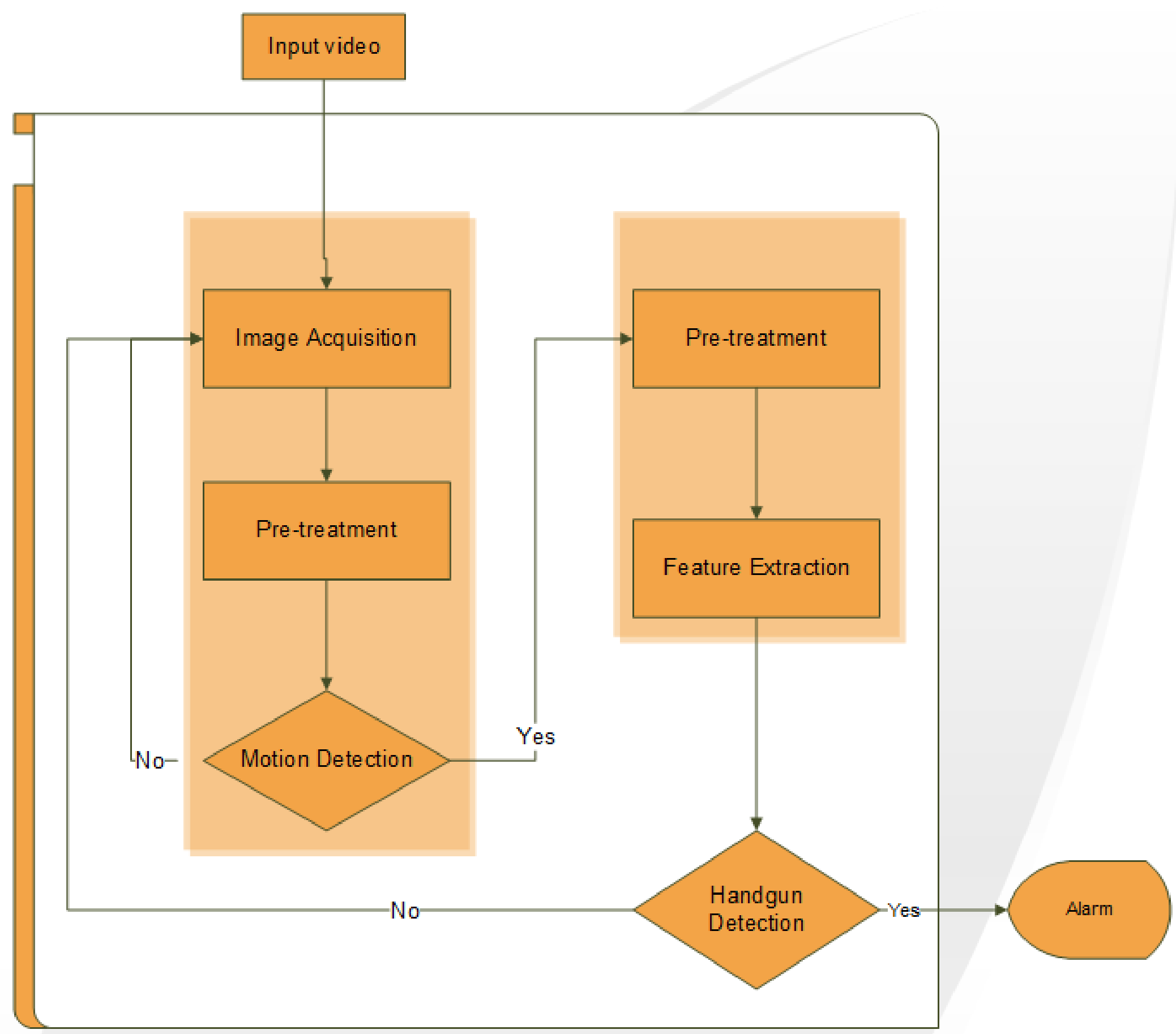

**Figure 2:** General block diagram of the proposed model

*2.1.1 Conversion to Gray Scale and Noise Reduction*

Prior to conducting any operations on the captured images, it is crucial to convert them to grayscale. Utilizing grayscale images simplifies the process and is more efficient for the intended task. Furthermore, reducing the noise caused by cameras and lighting is essential. This can be accomplished by averaging each pixel with its neighboring pixels.

*2.1.2 Application of Threshold*

At this stage of the process, the aim is to transform the image into a binary format, which means having two possible pixel values. Pixels that surpass a certain threshold will be classified as white, while the remaining pixels will be classified as black. This binary representation helps in identifying the moving object within the image [16].

*2.1.3 Contour detection*

Once the region of interest (ROI) is obtained from the image, the subsequent step is to identify and detect the outlines within this specific region. If no contours are detected after this process, it signifies the absence of movement, and the system reverts to the image acquisition step. However, if contours are detected, it initiates the process of handgun detection [17].

*2.2 Handgun Detection*

The described process depends on a computer vision module that employs a CNN-based model [18] for handgun detection. The CNN comprises nodes, where an input tensor is fed and another tensor is produced as output from the final nodes. The input tensor represents the input image, while the output tensor indicates the binary classification label for detection or non-detection of handguns.

The detection of handguns involves exploring various solutions to address the challenge of real-time detection. However, one of the major hurdles lies at the hardware level. Consequently, experiments were conducted using four different models to address this issue:

1. Model based on CNN.
2. Model based on Fast R-CNN [5].
3. Model based on MobileNet [19].
4. Model based on AlexNet [20].

*2.2.1 CNN based Model*

The first model, depicted in Figure 3, consists of five convolution layers, two maximum pooling layers, and three fully connected layers. The input image has a size of 32x32 pixels. The image is first passed through the initial convolution layer, which comprises 32 filters of size 3x3. Each convolutional layer is followed by a rectified linear unit (ReLU) activation function, which ensures that the neurons only output positive values. This convolution operation generates 32 feature maps of size 32x32.

The resulting features are then fed into the second convolution layer, which consists of 32 filters. After the convolution, a ReLU activation function is once again applied, and then a pooling operation that lowers the size of the feature maps and the quantity of parameters follows. The output of this layer is 32 feature maps of size 16x16. This process is repeated for the third, fourth, and fifth convolutional layers, with each layer having 64 filters. The ReLU activation function is applied after each operation of convolution.
Following the five convolutional layers, a pooling layer is employed, which produces 64 feature maps with dimensions of 8x8. The feature vector extracted from these convolutions possesses a dimension of 4096.
Afterward, a neural network with three fully connected layers is utilized. The first two layers consist of 1024 neurons each, employing the ReLU activation function. The third layer uses a SoftMax activation function to compute the probability distribution across the 100 classes (corresponding to the number of classes in the Handgun Dataset for the sliding window approach).

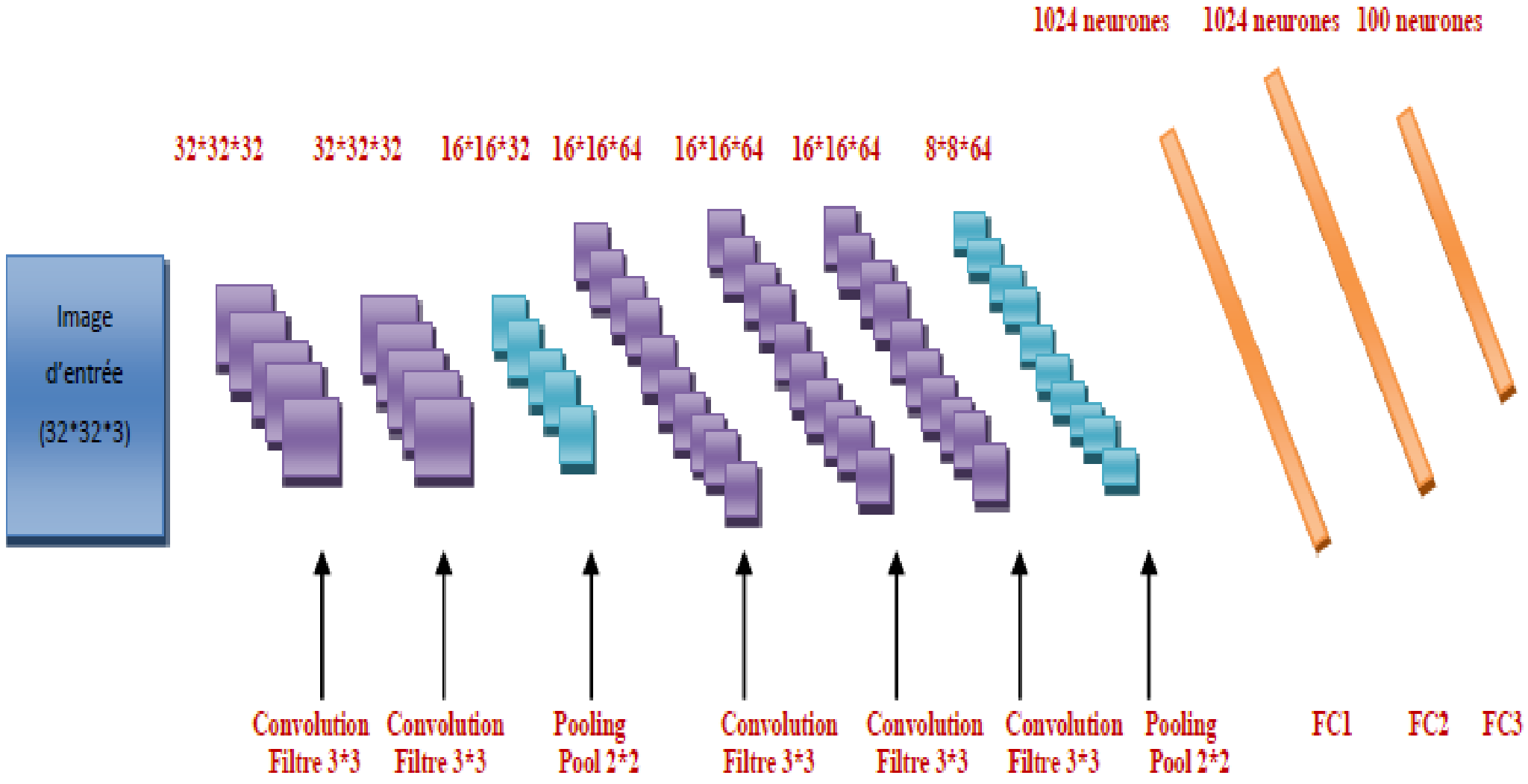

**Figure 3:** Proposed CNN Architecture [21]

*2.2.2 Fast R-CNN based Model*

The second model, illustrated in Figure 4, takes region proposals generated by an external system (such as a selective search) as input. Following the proposals, they undergo a pooling layer specifically designed for the region of interest (ROI). This pooling layer resizes each region and its corresponding data to a fixed size. This resizing step is essential because the fully connected layer requires all vectors to have the same size for further processing. By resizing the regions, consistency in the input dimensions is ensured for further processing in the model.

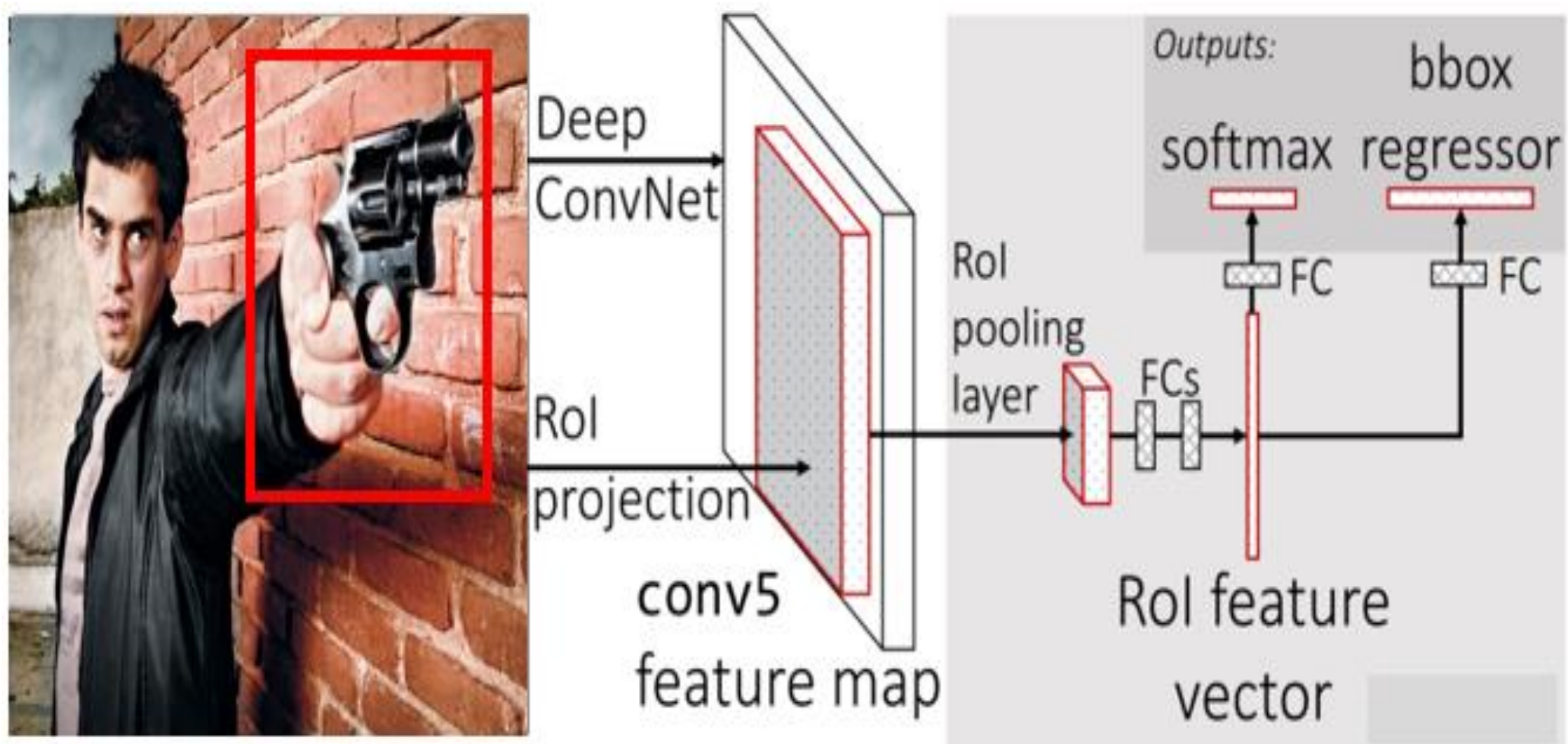

**Figure 2:** Fast R-CNN architecture [5]

*2.2.3 MobileNet based Model*

This model shares a similar architecture with the second model, with slight adjustments made to the configuration of learning parameters. These adjustments might include changes in the learning rate, regularization techniques, or optimization algorithms used during the training process. These modifications aim to improve the model's performance and adapt it to specific requirements or challenges encountered in the task of handgun detection.

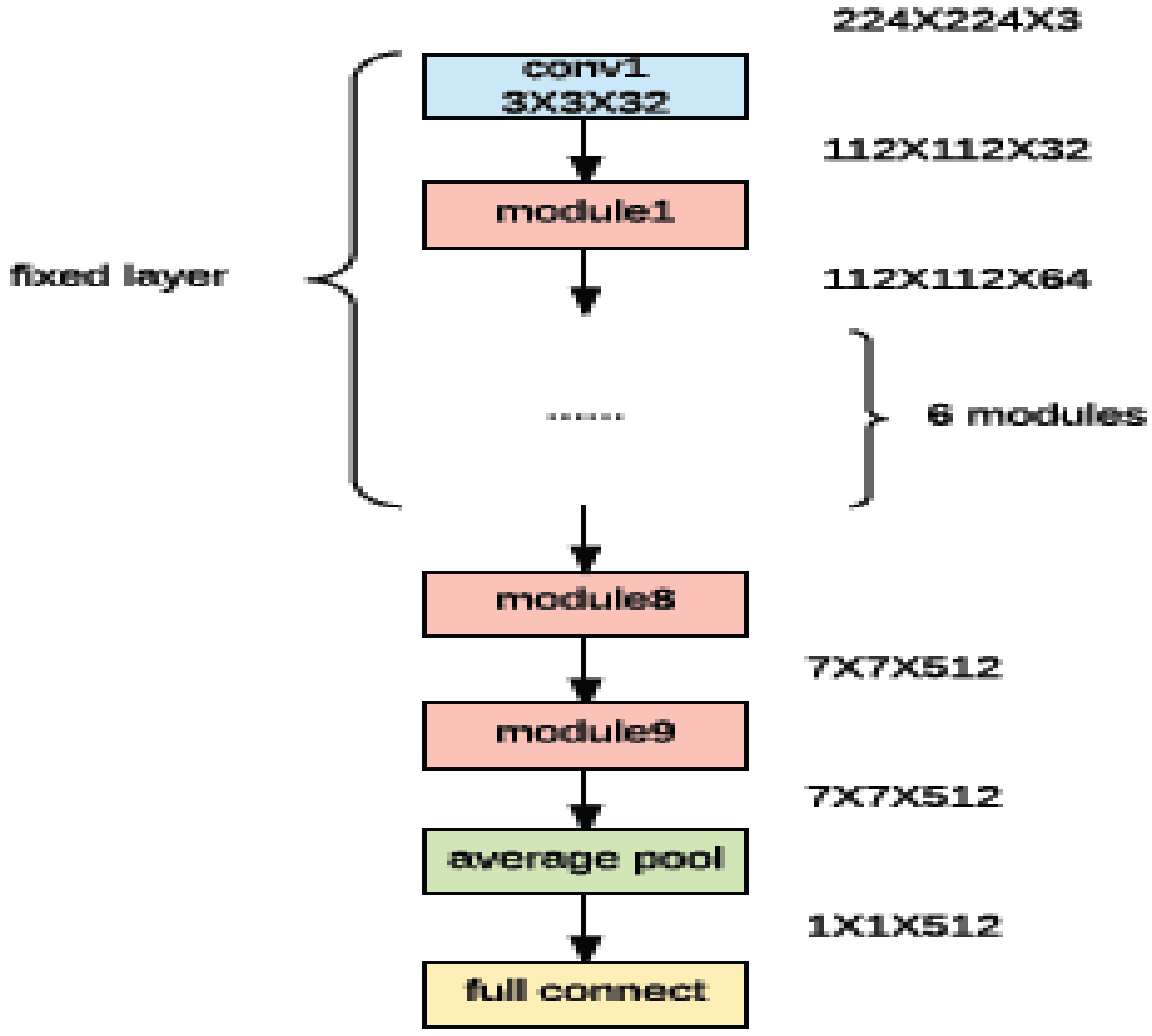

**Figure 3:** MobileNet architecture [21]

*2.2.4 AlexNet based Model*

To expedite the learning task and reduce the required time and epochs, it is possible to retrain a pre-trained image classification network. In the same manner that people employ their prior knowledge to understand and accomplish new problems, neural networks are trained and tested on various datasets [22]. This pre-trained network has already learned to extract powerful and informative features from natural images. By using this pre-trained network as a starting point, it becomes easier to learn a new task efficiently.

AlexNet is indeed one of these pre-trained networks. It is a convolutional neural network that has already been trained on over a million images from the ImageNet database [23]. The AlexNet network is 8 layers deep and has the capability to classify images into 1000 object categories, including items like keyboards, mice, pencils, and various animals. This extensive training has enabled the network to learn complex feature representations for a diverse set of images. The network's image input size is 227 x 227 pixels.

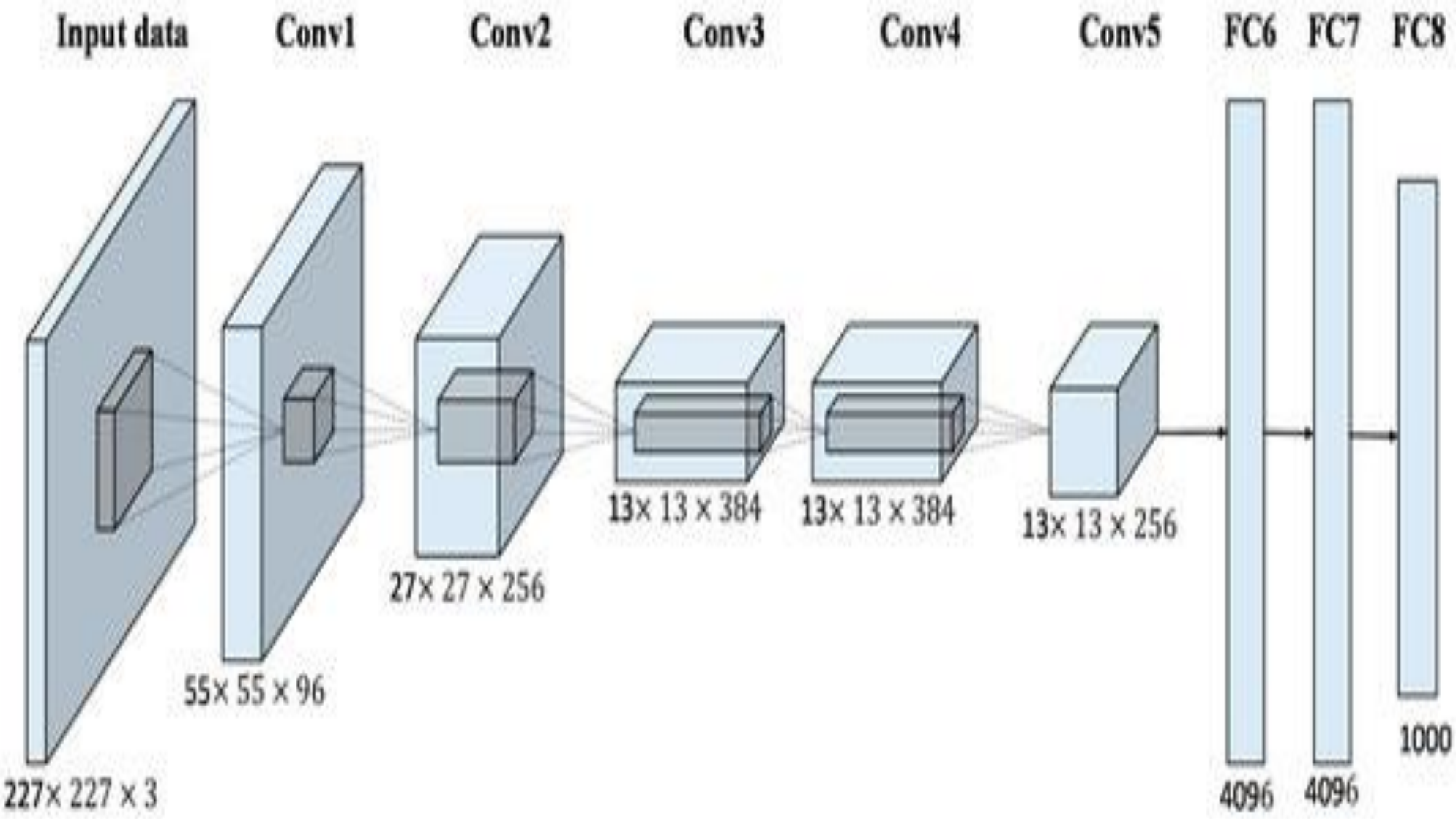

**Figure 4:** The AlexNet archit\ecture with spatial pyramid pooling [24]

## 3. Experiments

To assess the effectiveness of the proposed models, two databases [25] are utilized during the learning phase.

*3.1 Handgun Dataset for the sliding window approach.*

The training dataset used for the classification task consists of 102 classes with a total of 9,261 images. Among these classes, the handgun class contains 200 images.

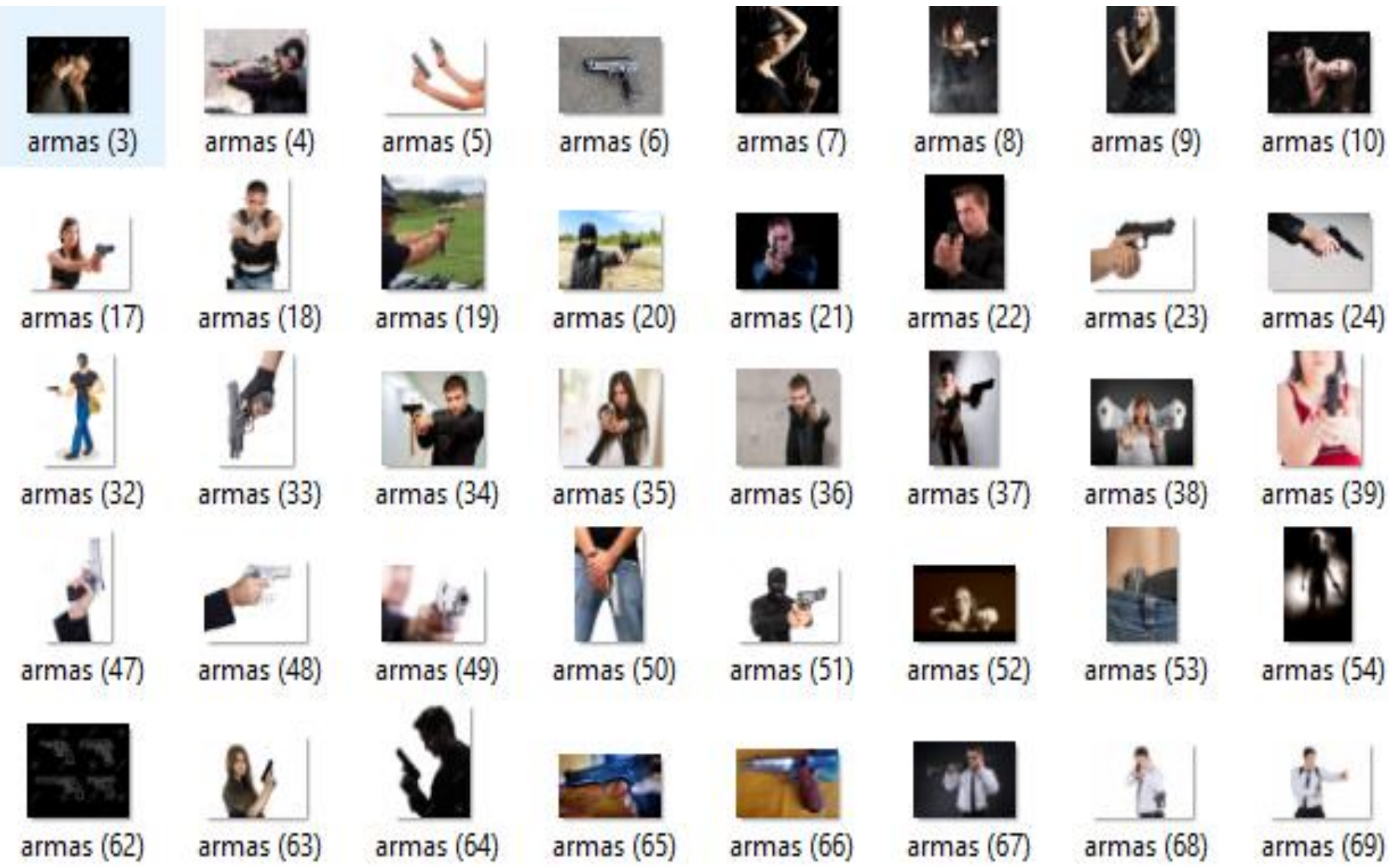

**Figure 7:** Samples from the Handgun Dataset for the Sliding Window Approach

*3.2 Handgun Dataset for the region.*

The training dataset used for the detection task comprises 3,000 images specifically focused on handguns. These images are meticulously chosen to encompass a comprehensive context surrounding the handguns.

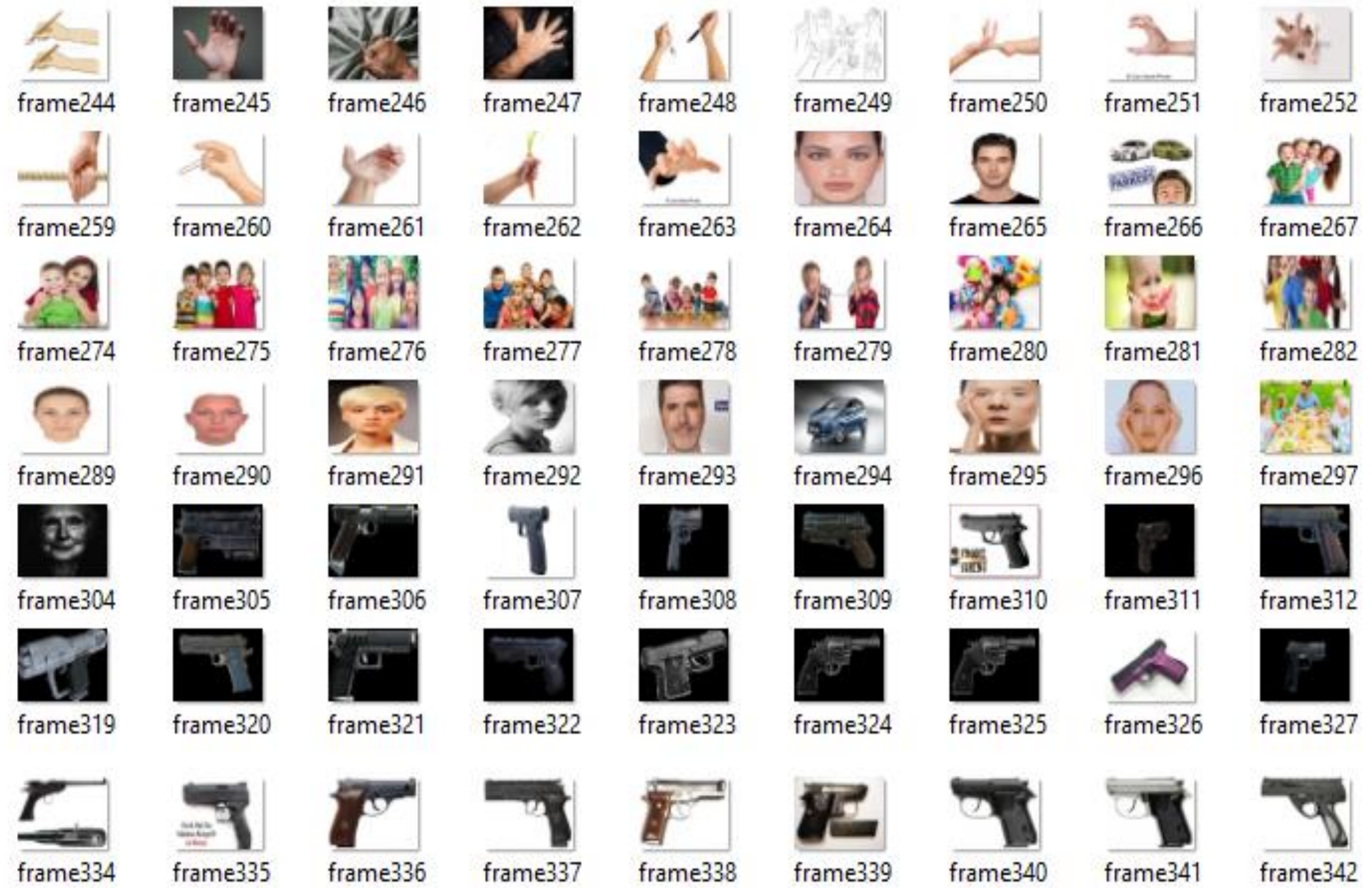

**Figure 8:** Samples from the Handgun Dataset for the Region Proposal Approach

The test dataset employed for both classification and detection tasks comprises a total of 608 images. Out of these, 304 images are specifically focused on handguns, while the remaining 304 images represent non-handgun objects.

For the first three models, the learning process involved utilizing a sample of 420 images from the Handgun Dataset database for the region proposal approach. The first and fourth models were then evaluated using the entire test dataset, which comprises 608 images (304 handgun images and 304 non-handgun images). The second model was tested with a subset of 200 images, while the third model was evaluated using a subset of 420 images.

## 4. Obtained results

In the evaluation of the models, the following metrics [26, 27, 28, 29, 30, 31, 32, 33, 34, 35] were used:

- Positive (P): The number of real positive cases in the dataset.
- Negative (N): The number of real negative cases in the dataset.
- True Positives (TP): This indicates the number of images where the process accurately detects a handgun among the 304 images that contain a handgun.
- True Negatives (TN): This refers to the number of images in which the process correctly does not detect a handgun among the 304 images that do not contain a handgun.
- False Positives (FP): This indicates the number of images in which the process incorrectly detects a handgun among the 304 images that do not contain a handgun.
- False Negatives (FN): This signifies the number of images in which the process fails to detect a handgun among the 304 images that contain a handgun.
- Accuracy (ACC): This is the degree of closeness of measurements of a quantity to that quantity's true value. It is calculated as (TP + TN) / (P + N).

$$\text{Accuracy} = \frac{\text{True Positives} + \text{True Negatives}}{\text{Positives} + \text{Negatives}} \quad (3)$$

- Precision (P): This represents the percentage of correctly detected handguns among the 304 images that contain handguns. It is calculated as TP / (TP + FP).

$$\text{Precision} = \frac{\text{True Positives}}{\text{True Positives} + \text{False Positives}} \quad (3)$$

- Recall (R): This indicates the percentage of correctly detected handguns among the entire 608 images in the test dataset. It is calculated as TP / (TP + FN).

$$\text{Recall} = \frac{\text{True Positives}}{\text{True Positives} + \text{False Negatives}} \quad (4)$$

- F1 measure: This is a metric that combines precision and recall into a single value to provide a balanced evaluation. It is calculated as 2 * (precision * recall) / (precision + recall).

$$\text{F1}\, measure = 2 * \frac{Precision * Recall}{Precision + Recall} \quad (5)$$

These metrics help assess the performance of the models in terms of their ability to accurately detect handguns in the test dataset and provide a comprehensive evaluation of their precision, recall, and overall effectiveness.

The results obtained from the experiments demonstrate that the performance improves as the neural network learns from the database, particularly with an increase in the time of learning. The size and quality of the learning database also play a crucial role in achieving better results. When analyzing the results obtained, reference is made to Table 2, where it is observed that the learning error and validation decrease as the learning time increases.
For the first model, a total of 274 images were misclassified. On the other hand, 334 images were correctly classified. As a result of limited learning time and hardware configuration constraints, the unsupervised classification model achieved an accuracy rate of 55%, indicating its performance is suboptimal.

In contrast, the second model achieved better results. It misclassified only 40 images. The remaining 160 images were correctly classified, yielding an accuracy rate of 80%. However, for real-time testing using a webcam, this model indeed demands a robust hardware configuration, particularly the computational capabilities provided by a GPU, to achieve optimal performance.

The third model exhibited superior performance, misclassifying only 42 images. A significant majority of the images (378) were correctly classified, resulting in an impressive accuracy rate of 90% for static images. This model also showed good performance in real-time testing. As depicted in Figure 6, compared to the other previous models, the proposed model requires less time for learning and performs better during execution (testing). Furthermore, this proposed transfer learning-based model exhibited good performance even when tested on larger datasets, achieving an accuracy rate of 86.68%.

**Table 2:** Comparison of all models

| Model | Number of Images | Number of Well Classified Images | Number of Misclassified Images | Learning Time (Hours) | Accuracy |
|---|---|---|---|---|---|
| CNN | 608 | 334 | 274 | 5 | 55 |
| Fast R-CNN | 200 | 160 | 40 | 24 | 80 |
| MobileNet | 420 | 378 | 42 | 192 | 90 |
| AlexNet | 608 | 527 | 81 | 9 | 86.68 |

Overall, the results highlight the importance of sufficient learning time, a large and diverse learning database, and powerful hardware configurations to achieve optimal performance in handgun detection tasks. The proposed model, utilizing a transfer learning approach, demonstrated the highest accuracy-to-learning time ratio.
Table 2 showcases the results obtained from the different models used in this study in terms of accuracy. The utilization of the fast R-CNN and MobileNet-CNN models involves initiating a fresh learning process using the Handgun Dataset, which comprises 3,000 images for region proposals. This learning process spans 48 hours. Additionally, the last classification layer of the proposed AlexNet-based model is modified to accommodate two classes instead of the original 1,000. It is then retrained using 1,000 non-handgun images from the Handgun Dataset for the sliding window approach and 1,000 handgun images from the Handgun Dataset for the region proposals approach [10]. The retraining process takes 9 hours to complete.

Furthermore, the evaluation is performed using the other metrics on a test set containing 608 images, consisting of 304 handgun images and 304 non-handgun images. As depicted in Table 3, all models exhibit good performance on static images and demonstrate real-time capabilities. However, the proposed AlexNet-based model stands out with a higher ratio of precision, recall, and F1 measure compared to the other trained models. It is important to note that, in comparison to other studies in the literature, more improvements can be achieved.

It is worth mentioning that all the learning tasks in this work were conducted on basic machines with modest hardware configurations. Despite the hardware limitations, the models were able to achieve satisfactory results, showcasing their potential for application in real-world scenarios.

**Table 3:** Obtained Results

| Model | TP | FN | TN | FP | P | R | F1 |
|---|---|---|---|---|---|---|---|
| Fast R-CNN | 232 | 72 | 248 | 56 | 80.76 | 76.31 | 78.37 |
| MobileNet | 156 | 54 | 168 | 42 | 78.78 | 74.28 | 76.46 |
| Alexnet | 272 | 32 | 255 | 49 | 84.74 | 89.47 | 87.04 |
| [10] | 304 | 0 | 247 | 57 | 84.21 | 100 | 91.43 |

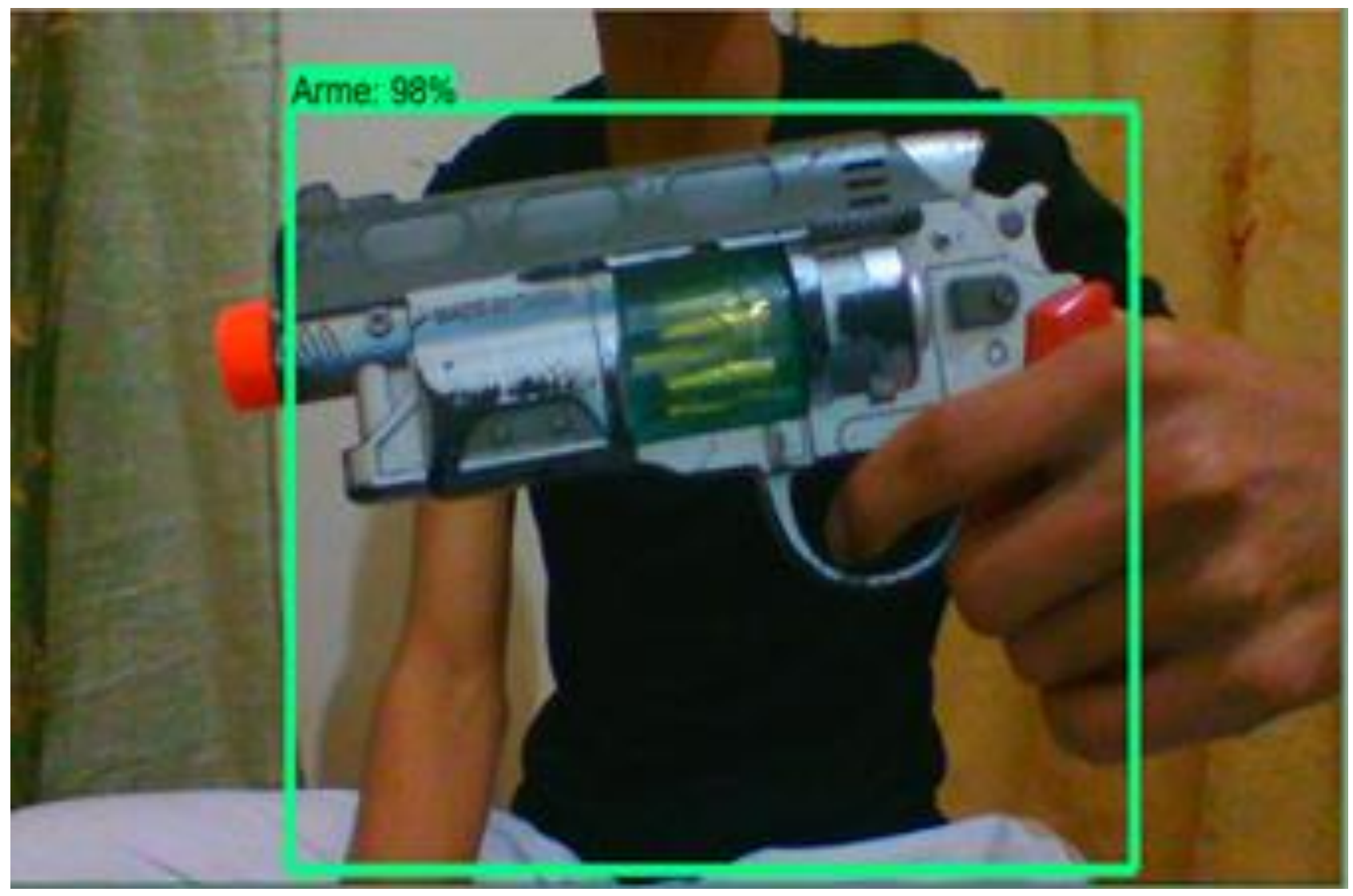

**Figure 9:** Detection of a handgun in real-time in an image captured by a webcam

## 5. Conclusion

In conclusion, this study significantly contributes to the field by highlighting the potential of AI as an efficient solution for real-time surveillance, thereby enhancing security during significant events. The research's primary components, motion detection and real-time handgun detection, have been thoroughly explored. While implementing deep learning for handgun detection presented challenges, particularly due to computational requirements, the study demonstrates the feasibility of utilizing soft deep learning models for handgun detection, albeit with longer learning times, even when constrained by a CPU. Furthermore, the results obtained from the proposed-based model emphasize its potential to reduce learning time and overcome hardware limitations by leveraging pre-trained networks such as AlexNet while maintaining detection performance. As a future direction, additional experimentation can be conducted on the proposed transfer learning model, utilizing more powerful machines to further optimize the learning process and enhance the handgun detection system's overall performance, thus advancing the existing literature in the field of AI-based surveillance.